\documentclass[11pt]{article}
\usepackage{acl2001,times,epsfig}
\title{Combining a self-organising map with memory-based learning}
\author{James Hammerton \\
	james.hammerton@ucd.ie \\ 
	Dept of Computer Science \\
        University College Dublin \\
        Ireland \\
	\And 
        Erik F. Tjong Kim Sang \\	
        erikt@uia.ua.ac.be  \\
	CNTS -- Language Technology Group \\
	University of Antwerp \\
	Belgium}
\date{}
\begin{document}
\maketitle

\begin{abstract}
Memory-based learning (MBL) has enjoyed considerable success in
corpus-based natural language processing (NLP) tasks and is thus a
reliable method of getting a high-level of performance when building
corpus-based NLP systems. However there is a bottleneck in MBL whereby
any novel testing item has to be compared against all the training
items in memory base. For this reason there has been some interest in
various forms of memory editing whereby some method of selecting a
subset of the memory base is employed to reduce the number of
comparisons. This paper investigates the use of a modified
self-organising map (SOM) to select a subset of the memory items for
comparison. This method involves reducing the number of comparisons to
a value proportional to the square root of the number of training
items. The method is tested on the identification of base noun-phrases
in the Wall Street Journal corpus, using sections
15 to 18 for training and section 20 for testing.
\end{abstract}

\bibliographystyle{acl}

\section{Introduction}
\label{Intro}

Currently, there is considerable interest in machine learning methods
for corpus-based language learning. A promising technique here is
memory-based learning\footnote{Also known as instance based
learning.} (MBL) \cite{cnlp:daelemans99}, where a task is redescribed
as a classification problem. The classification is performed by
matching an input item to the most similar of a set of training items
and choosing the most frequent classification of the closest
item(s). Similarity is computed using an explicit similarity metric.

MBL performs well by bringing all the training data to bear on the
task. This is done at the cost, in the worst case, of comparing novel
items to all of the training items to find the closest match. There is
thus some interest in developing memory editing techniques to select a
subset of the items for comparison.

This paper investigates whether a self-organising map (SOM) can be
used to perform memory editing without reducing performance. The
system is tested on base noun-phrase (NP) chunking using the Wall
Street Journal corpus \cite{cnlp:marcus93}.

\section{The Self-Organising Map (SOM)}

\begin{figure}
\setlength{\epsfxsize}{1.5in}
\setlength{\epsfysize}{2in}
\centerline{\epsffile{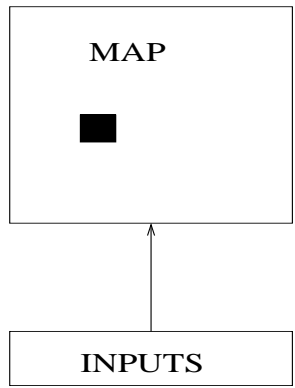}}
\caption{The Self-Organising Map. The map units respond to the
inputs. The map unit whose weight vector is closest to the input
vector becomes the winner. During training, after a winner is chosen,
the weight vectors of the winner and a neighbourhood of surrounding
units are nudged towards the current input. }
\label{SOM}
\end{figure}

The SOM was developed originally by Kohonen \shortcite{cnlp:kohonen90} and
has found a wide range of uses from classification to storing a
lexicon. It operates as follows (see Figure~\ref{SOM}). The SOM
consists of two layers, an input layer and the map. Each unit in the
map has a vector of weights associated with it that is the same size
as that of the input layer. When an input is presented to the SOM, the
unit whose weight vector is closest to the input vector is selected as
a winner.

During training, the weight vectors of winning unit and a set of units
within the neighbourhood of the winner are nudged, by an amount
determined by the learning rate, towards the input vector. Over time
the size of the neighbourhood is decreased. Sometimes the learning
rate may be too. At the end of training, the units form a map of the
input space that reflects how the input space was sampled in the
training data. In particular areas of input space in which there were
a lot of inputs will be mapped in finer detail, using more units than
areas where the inputs were sparsely distributed.

\section{Why use a SOM for memory editing}

The SOM was chosen because the input is matched to the unit with the
closest weight vector. Thus it is motivated by the same principle as
used to find the closest match in MBL. It is thus hoped that the SOM
can minimise the risk of failing to select the closest match, since
the subset will be chosen according to similarity.

However, Daelemans {\em et al} \shortcite{cnlp:daelemans99b} claim that, in
language learning, pruning the training items is harmful. When they
removed items from the memory base on the basis of their typicality
(i.e. the extent to which the items were representative of other items
belonging to the same class) or their class prediction strength
(i.e. the extent to which the item formed a predictor for its class),
the generalisation performance of the MBL system dropped across a range
of language learning tasks.

The memory editing approach used by Daelemans {\em et al} removes
training items independently of the novel items, and the remainder are
used for matching with all novel items. If one selects a {\em
different} subset for each novel item based on similarity to the novel
item, then maybe the risk of degrading performance in memory editing
will be reduced. This work aims to achieve precisely this.

\section{A hybrid SOM/MBL classifier}

\subsection{Labelled SOM and MBL (LSOMMBL)}
\label{lsommbl}
A modified SOM was developed called Labelled SOM. Training proceeds as
follows:

\begin{itemize}
\item Each training item has an associated label. Initially all the
map units are unlabelled.

\item When an item is presented, the closest unit out of those with
the same label as the input and those that are unlabelled is chosen as
the winner. Should an unlabelled unit be chosen it gets labelled with
the input's label.

\item The weights for neighbouring units are updated as with the
standard SOM if they share the same label as the input or are unlabelled.

\item When training ends, all the training inputs are presented to the
SOM and the winners for each training input noted. Unused units are
discarded.
\end{itemize}

\noindent Testing proceeds as follows:

\begin{itemize}
\item When an input is presented a winning unit is found {\em for each
category}.

\item The closest match is selected from the training items associated
with each of the winning units found. 

\item The most frequent classification for that match is
chosen. 

\end{itemize}

\noindent It is thus hoped that the closest matches for each category
are found and that these will include the closest match amongst them.

Assuming each unit is equally likely to be chosen, the average number
of comparisons here is given by $C(N + I/N)$ where $C$ is the number of
categories, $N$ is the number of units in the map and $I$ is the number of
training items. Choosing $N=\sqrt{I}$ minimises comparisons to
$2C\sqrt{I}$. In the experiments the size of the map was chosen to be
close to $\sqrt{I}$. This system is referred to as LSOMMBL.

\subsection{SOM and MBL (SOMMBL)}
In the experiments, a comparison with using the standard SOM in a
similar manner was performed. Here the SOM is trained as normal on
the training items. 

At the end of training each item is presented to the SOM and the
winning unit noted as with the modified SOM above. Unused units are
discarded as above. 

During testing, a novel item is presented to the SOM and the top C
winners chosen (i.e. the $C$ closest map units), where C is the number
of categories. The items associated with these winners are then
compared with the novel item and the closest match found and then the
most frequent classification of that match is taken as before. This
system is referred to as SOMMBL.

\section{The task: Base NP chunking}

The task is base NP chunking on section 20 of the Wall Street Journal
corpus, using sections 15 to 18 of the corpus as training data as in
\cite{cnlp:ramshaw95}. For each word in a sentence, the POS tag is
presented to the system which outputs whether the word is inside or
outside a base NP, or on the boundary between 2 base NPs.

Training items consist of the part of speech (POS) tag for the current
word, varying amounts of left and right context (POS tags only) and
the classification frequencies for that combination of tags. The tags
were represented by a set of vectors. 2 sets of vectors were used for
comparison. One was an orthogonal set with a vector of all zeroes for
the ``empty'' tag, used where the context extends beyond the
beginning/end of a sentence. The other was a set of 25 dimensional
vectors based on a representation of the words encoding the contexts
in which each word appears in the WSJ corpus. The tag representations
were obtained by averaging the representations for the words appearing
with each tag. Details of the method used to generate the tag
representations, known as lexical space, can be found in
\cite{cnlp:zavrel96}. Reilly \shortcite{cnlp:reilly98} found it beneficial when
training a simple recurrent network on word prediction.

The self-organising maps were trained as follows:

\begin{itemize}

\item For maps with 100 or
more units, the training lasted 250 iterations and the neighbourhood
started at a radius of 4 units, reducing by 1 unit every 50 iterations
to 0 (i.e. where only the winner's weights are modified). The learning
rate was constant at 0.1.

\item For the maps with 6 units, the training lasted 90 iterations, with an
initial neighbourhood of 2, reducing by one every thirty
iterations. 

\item For the maps with 30 units, the training lasted 150 iterations, with
an initial neighbourhood of 2, reduced by one every 50 iterations. A
single training run is reported for each network since the results did
not vary significantly for different runs.

\end{itemize}

These map sizes were chosen to be close to the square root of the
number of items in the training set. No attempt was made to
systematically investigate whether these sizes would optimise the
performance of the system. They were chosen purely to minimise the
number of comparisons performed.

\begin{table*}[t]
\centering
\begin{tabular}{|l|l|l|l|l|l|}
\hline
 	& features & window 	& Chunk  & Chunk tag & Max \\
	&          &            & fscore & accuracy  & comparisons (\% of items)\\
\hline
LSOMMBL & lex	  & 0-0 	& 79.99 & 94.48\% & 87 (197.7\%) \\
LSOMMBL & lex	  & 1-0		& 86.13 & 95.77\% & 312 (30.3\%)\\
LSOMMBL & lex	  & 1-1		& {\bf 89.51} & 96.76\% & 2046 (20.4\%)\\
LSOMMBL & lex	  & 2-1		& 88.91 & 96.42\% & 2613 (6.8\%)\\
\hline
LSOMMBL & orth 	  & 0-0		& 79.99 & 94.48\% & 87 (197.7\%)\\
LSOMMBL & orth	  & 1-0		& 86.09 & 95.76\% & 702 (68.1\%)\\
LSOMMBL & orth	  & 1-1		& 89.39 & 96.75\% & 1917 (19.1\%)\\
LSOMMBL & orth	  & 2-1		& 88.71 & 96.52\% & 2964 (7.7\%) \\
\hline
SOMMBL	& lex	  & 0-0		& 79.99 & 94.48\% & 51 (115.9\%)\\
SOMMBL	& lex	  & 1-0		& 86.11 & 95.77\% & 327 (31.7\%) \\
SOMMBL	& lex	  & 1-1		& {\bf 89.47} & 96.74\% & 1005 (10.0\%) \\
SOMMBL	& lex	  & 2-1		& 88.98 & 96.48\% & 1965 (5.1\%) \\
\hline
SOMMBL	& orth	  & 0-0		& 79.99 & 94.48\% & 42 (95.5\%) \\
SOMMBL	& orth	  & 1-0		& 86.08 & 95.75\% & 306 (29.7\%) \\
SOMMBL	& orth	  & 1-1		& 89.38 & 96.77\% & 1365 (13.6\%) \\
SOMMBL	& orth	  & 2-1		& 88.61 & 96.45\% & 2361 (6.1\%)\\
\hline
MBL 	& tags	  & 0-0		& 79.99 & 94.48\% & 44 (100.0\%) \\
MBL 	& tags	  & 1-0		& 86.14 & 95.78\% & 1031 (100.0\%) \\
MBL 	& tags	  & 1-1		& 89.57 & 96.80\% & 10042 (100.0\%)\\
MBL 	& tags	  & 2-1		& 89.81 & 96.83\% & 38465 (100.0\%)\\
\hline
MBL	& lex 	  & 0-0		& 79.99 & 94.70\% & 44 (100.0\%)\\
MBL 	& lex 	  & 1-0		& 86.14 & 95.95\% & 1031 (100.0\%) \\
MBL 	& lex 	  & 1-1		& 89.57 & 96.93\% & 10042 (100.0\%)\\
MBL 	& lex 	  & 2-1		& {\bf 89.81} & 96.96\% & 38465 (100.0\%)\\
\hline 
MBL	& orth	  & 0-0		& 79.99 & 94.70\% & 44 (100.0\%)\\
MBL	& orth	  & 1-0		& 86.12 & 95.94\% & 1031 (100.0\%) \\
MBL	& orth	  & 1-1		& 89.46 & 96.89\% & 10042 (100.0\%)\\
MBL	& orth	  & 2-1		& 89.55 & 96.87\% & 38465 (100.0\%)\\
\hline
\end{tabular}
\caption{Results of base NP chunking for section 20 of the
WSJ corpus, using SOMMBL, LSOMMBL and MBL, training was performed on
sections 15 to 18. The fscores of the best performers in each case for
LSOMMBL, SOMMBL and MBL have been highlighted in {\bf bold}. 
%Note that
%the chunk tag accuracy is not a good measure of performance for
%natural language tasks and is included purely for comparison with
%\cite{cnlp:daelemans99b}.
}
\label{Results}
\end{table*}

\section{Results}

Table~\ref{Results} gives the results of the experiments. The columns
are as follows:

\begin{itemize}
\item ``features''. This column indicates how the features are made
up. 
\\ ``lex'' means the features are the lexical space vectors
representing the POS tags. ``orth'' means that orthogonal vectors are
used. ``tags'' indicates that the POS tags themselves are used.
% modified by ET
MBL uses a weighted overlap similarity metric while SOMMBL and LSOMMBL
use the euclidean distance.
% and the
%number of mismatches between the tags in the test item and those in
%the training item is the similarity metric. Otherwise, the
%similarity metric is the euclidean distance.

\item ``window''. This indicates the amount of context, in the form
of ``left-right'' where ``left'' is the number of words in the left
context, and ``right'' is the number of words in the right context. 

\item ``Chunk fscore'' is the fscore for finding base NPs. The fscore ($F$)
is computed as $F=\frac{2PR}{P+R}$ where $P$ is the percentage of base NPs
found that are correct and $R$ is the percentage of base NPs defined in
the corpus that were found.

\item ``Chunk tag accuracy'' gives the percentage of correct chunk tag
classifications. This is provided to give a more direct comparison
with the results in \cite{cnlp:daelemans99}. However, for many NL
tasks it is not a good measure and the fscore is more
accurate.

\item ``Max comparisons''. This is the maximum number of comparisons
per novel item computed as $C(N+X)$ where $C$ is the number of
categories, $N$ is the number of units and $X$ is the maximum number
of items associated with a unit in the SOM. This number depends on how
the map has organised itself in training. The number given in brackets
here is the percentage this number represents of the maximum number of
comparisons under MBL. The average number of
comparisons is likely to be closer to the average mentioned in
Section~\ref{lsommbl}.
\end{itemize}

\begin{table}[t]
\centering
\begin{tabular}{|l|l|l|}
\hline
window  & SOM   & Training \\
        & size  & items    \\
\hline
0-0     & 10    & 44       \\
1-0     & 30    & 1131     \\
1-1     & 100   & 10042    \\
2-1     & 200   & 38465    \\
\hline
\end{tabular}
\caption{Network sizes and number of training items}
\label{sizes}
\end{table}

Table~\ref{sizes} gives the sizes of the SOMs used for each context
size, and the number of training items.

For small context sizes, LSOMMBL and MBL give the same performance. As
the window size increases, LSOMMBL falls behind MBL. The worst drop in
performance is just over 1.0\% on the fscores (and just over 0.5\% on
chunk tag accuracy). This is small considering that e.g. for the
largest context the number of comparisons used was at most 6.8\% of
the number of training items.

To investigate whether the method is less risky than the memory
editing techniques used in \cite{cnlp:daelemans99b}, a re-analysis of
their data in performing the same task, albeit with lexical
information, was performed to find out the exact drop in the chunk tag
accuracy.

In the best case with the editing techniques used by
\cite{cnlp:daelemans99b}, the drop in performance was 0.66\% in the
chunk tag accuracy with 50\% usage (i.e. 50\% of the training items
were used). Our best involves a drop of 0.23\% in chunk tag accuracy
and only 20.4\% usage. Furthermore their worst case involves a drop of
16.06\% in chunk tag accuracy again at the 50\% usage level, where
ours involves only a 0.54\% drop in accuracy at the 6.8\% usage
level. This confirms that our method may be less risky, although a
more direct comparison is required to demonstrate this in a fully
systematic manner. For example, our system does not use lexical
information in the input where theirs does, which might make a
difference to these results.

Comparing the SOMMBL results with the LSOMMBL results for the same
context size and tagset, the differences in performance are insignificant,
typically under 0.1 points on the fscore and under 0.1\% on the chunk
tag accuracy. Furthermore the differences sometimes are in favour of
LSOMMBL and sometimes not. This suggests they may be due to noise due
to different weight initialisations in training rather than a
systematic difference. 

Thus at the moment it is unclear whether SOMMBL or LSOMMBL have an
advantage compared to each other. It does appear however that the use
of the orthogonal vectors to represent the tags leads to slightly
worse performance than the use of the vectors derived from the lexical
space representations of the words.

\section{Discussion and future work}

The results suggest that the hybrid system presented here is capable
of significantly reducing the number of comparisons made without
risking a serious deterioration in performance. Certainly, the
reduction in the number of comparisons was far greater than with the
sampling techniques used by Daelemans {\em et al} 
while yielding similar levels of deterioration.
%and yielded  similar levels of deterioration.

Given that a systematic investigation into what the optimal training
regimes and network sizes are has not been performed the results thus
seem promising.

With regard to the network sizes, for example, one reviewer commented
that choosing the number of units to be the square root of the number
of training items may not be optimal since the clusters formed in
self-organising maps theoretically vary as the squared cube root of
the density, thus implying that a larger number of units may offer
better performance. What impact using this insight would have on the
performance (as opposed to the speed) of the system however is
unclear. Another variable that has not been systematically
investigated is the optimal number of winning units to choose during
testing. Increasing this value should allow tuning of the system to
balance deterioration of the performance against the reduction in
comparisons made.

Another issue regards the nature of the comparisons made. When
choosing a winning unit, the input is compared to the centroid of the
cluster each unit represents. However this fails to take into account
the distribution of the items around the centroid. As one reviewer
suggested, it may be that if instead the comparison is made with the
periphery of the cluster there would be a reduced risk of missing the
closest item. One possibility for taking this into account would be to
compute the average distance of the items from the centroid and
subtract this from the raw distance computed between the input and the
centroid. This would take into account how spread out the cluster is
as well as how far the centroid is from the input item. This would be
a somewhat more expensive comparison to make but may be worthwhile to
improve the probability that the closest item is found in the clusters
that are searched.

Future work will therefore include systematically investigating these
issues to determine the optimal parameter settings. Also a more
systematic comparison with other sampling techniques, especially other
methods of clustering, is needed to confirm that this method is less
risky than other techniques.

\section{Conclusion}

This work suggests that using the SOM for memory editing in MBL may be
a useful technique for improving the speed of MBL systems whilst
minimising reductions in performance. Further work is needed however to
find the optimal training parameters for the system and to confirm the
utility of this approach.

\section*{Acknowledgements}

The authors would like to thank the following people for their
discussions and advice during this work; the reviewers of this paper
for some helpful comments, Antal van den Bosch for providing the data
from \cite{cnlp:daelemans99b} for our analysis, Ronan Reilly for
his discussions of this work, Walter Daelemans \& other members of the
CNTS who made the first author's visit to CNTS an enjoyable and
productive stay.

This work was supported by the EU-TMR project ``Learning Computational
Grammars''. The (L)SOMMBL simulations were performed using
the PDP++ simulator. The MBL simulations were performed using the
TiMBL package

\bibliographystyle{acl}

\end{document}